# An NLP Solution to Foster the Use of Information in Electronic Health Records for Efficiency in Decision-Making in Hospital Care

**Abstract.** The project aimed to define the rules and develop a technological solution to automatically identify a set of attributes within free-text clinical records written in Portuguese. The first application developed and implemented on this basis was a structured summary of a patient's clinical history, including previous diagnoses and procedures, usual medication, and relevant characteristics or conditions for clinical decisions, such as allergies, being under anticoagulant therapy, etc. The project's goal was achieved by a multidisciplinary team that included clinicians, epidemiologists, computational linguists, machine learning researchers and software engineers, bringing together the expertise and perspectives of a public hospital, the university and the private sector. Relevant benefits to users and patients are related with facilitated access to the patient's history, which translates into exhaustiveness in apprehending the patient's clinical past and efficiency due to time saving.

Adelino Leite-Moreira[1,2], Afonso Mendes[3], Afonso Pedrosa[1], Amândio Rocha-Sousa[1,2], Ana Azevedo[1,2,4], André Amaral-Gomes[1,2], Cláudia Pinto[3], Helena Figueira[3], Nuno Rocha Pereira[1,2], Pedro Mendes[3] and Tiago Pimenta[1,2]

[1] Centro Hospitalar Universitário de São João. Porto, Portugal
{jose.pedrosa, a.oliveira, andre.gomes, nmiguel, joao.pimenta}@chsj.min-saude.pt
[2] Faculdade de Medicina da Universidade do Porto. Porto, Portugal
{amoreira, arsousa}@med.up.pt
[3] Priberam. Lisboa, Portugal
{amm, claudia.pinto, helena.figueira, pedro.mendes}@priberam.pt
[4] Instituto de Saúde Pública da Universidade do Porto. Porto, Portugal

## 1   Introduction

With the increasing complexity in patient disease trajectories, clinicians are often met with complex patient histories on top of which clinical decisions must be made. The societal expectations regarding effective benefit from existing healthcare resources, together with an increasing rate of adverse events and lower admissibility of error [1], support the need to enforce evidence-led medical decision-making using available health care data.

The end of the 20th century has brought enormous expectations that the use of electronic health records (EHR) would be a *panacea* to tackle problems of clinical effectiveness and particularly patient safety. However, more than 20 years later, despite the widespread use of EHR in healthcare settings, those objectives are far from attained. While legibility and access have improved immensely, problems of completeness, of balance between useful information and noise, and of guaranteeing that registered important clinical information is not disregarded by other health professionals persist. Also, the use of information for managerial/governance analytic purposes is limited by the enormous amount of information that is unstructured.

In another dimension, biomedical research follows a new model, with precision medicine approaches calling for the need and opportunity to study how data from both observational studies and clinical trials can be used to improve health outcomes in real-world practice settings, while fostering the use of clinical encounters data to learn and improve quality [2].

Requiring primary registration in structured formats by health professionals during clinical practice works well for very objective data, but it is unwelcomed to replace the description of history, signs, and symptoms. It also takes longer to try to fit more complex constructs into structured catalogs than to describe them narratively, leading clinicians to complain about spending more time doing registries than with the patient. Finally, registering in structured formats may become an obstacle to record aspects whose clinical interpretation is still uncertain, which are important for effective communication within the team, and a goldmine for data-driven discovery [3]. Solutions that allow the analysis of text contexts in the EHR can help to fill these gaps [4].

We are reporting the work undertaken by a multidisciplinary team that included clinicians, epidemiologists, computational linguists and information technology technicians, bringing together the expertise and perspectives of a public hospital, the university and the private sector. The setting is a public-, university-, tertiary-care-hospital, with in-patient and out-patient areas, emergency, intensive care, all medical and surgical specialties, for adults and children. Medication prescription and administration, laboratorial results, vital signs and a few scales are registered in a structured format. Nursing registries are mostly structured, based on the International Classification for Nursing Practice (ICNP®) [5]. Medical registries, including the clinical history at admission, daily clinical notes, surgical and imaging studies' reports, and discharge notes, are written in electronic support but as unstructured text. Hospitalization episodes are coded according to the International Classification of Diseases 10 [6] by trained dedicated medical professionals after discharge. The fact that the hospital has its EHR, J One, developed internally, facilitated the adaptation of the interface to the project's objectives, both in terms of usability and in terms of integration and information flow. This natural language processing (NLP) solution is prepared to integrate with other EHR applications.

As a contribution to the community, we are making the ontology and the annotation guidelines publicly available. In doing so, we hope to encourage other researchers and end-users to contribute with relevant extensions and datasets, enabling further and comparable research in Portuguese clinical NLP.

## 2    Ontology

The initial general objective was to provide a facilitated access to the patient's history, but we also envisioned from the beginning the improvement in diagnosis of diseases and incident adverse events during healthcare in real-time, as well as factors considered important in modifying clinical decisions.

For this, we needed an ontology that allowed: 1) to complete structured data (for example: diagnoses, procedures, medication, vital signs, laboratorial results, etc.), 2) to suggest structured data (diagnoses, procedures), 3) to characterize structured data (for example: regarding time, active or past, incident or prevalent) and 4) to give structured and controlled terminology to unstructured data (for example: functional status, signs/symptoms, behaviors and lifestyles, phenotypes).

When discussing the structure to systematize the constructs to capture from the text in the EHR, the clinical partners brought an emphatic apology in favor of using terms that are familiar to clinicians and used in daily clinical practice. This relates not only to the lexicon, but also to the level of disaggregation towards a relevant discriminatory capacity.

For diseases/diagnoses, limitations of statistical classifications such as the International Classification of Diseases for diagnoses are widely recognized and easy to understand when considering the purpose for which it was developed. A strong candidate that we considered was SNOMED CT [7], as it is a previously established and widely used ontology that covers a wide scope array of concepts and categories of data, with a more complex underlying structure, namely the consideration of the same term in several types of criteria (for example, the type of exam, the anatomic part/biological specimen on which it is made, etc.). However, it is not freely available, and it requires considerable training efforts to be used.

Therefore, considering clinical concern, group support and feasibility, we built an ontology with several classes and a hierarchy of up to 3 levels in each class. The classes were: "pathological conditions" [classified by organ system (cardiovascular, respiratory, neurological, etc.) and etiopathogeny (oncological, infectious, degenerative, etc.) with the same condition being repeated (for example, lung cancer is both oncological and respiratory)], "devices", "interventions" [second level: "surgeries", "medication" (chronic or transient, active or past), "chemotherapy", "radiotherapy", "physiotherapy", "ventilatory support", "renal replacement therapy", among others], "clinical findings (second level: "symptoms/signs" and "test results"), "anatomic structure", "gynecological/obstetric history" and "tests".

An important aspect we had to consider while structuring the ontology was the possibility for nesting or overlapping classes. This means that for annotation purposes all classes in the ontology can be connected among them by nesting. For example, in a phrase like "pleural effusion", which would belong to the "clinical findings" class, "pleural" would belong to the "anatomic structure" class. This nesting process is important for representing relationships between classes.

For each one of these classes, we also considered having modifier tags, which introduce information that would otherwise be lost by the simple use of the above-mentioned classes. For instance, modifiers like "negation", which is of the utmost importance for classifying constructs with particles that indicate the absence or the non-occurrence of some event or condition (e.g., "no food allergies), or "plan", which can be added to a class that describes interventions when they did not occur at the time when the EHR was written, but are planned to occur in the future (e.g., "x-ray scheduled for tomorrow"). The other modifiers we considered would add important information to the ontology classes were the following: "acute", "chronic", "worsened", "probable/possible", "normal", "augmented", "diminished", "beginning", "suspension", "ongoing" and "past". Some of these modifiers are common to all categories of the ontology (like "negation"), while others are only meant to be used with specific nodes (for instance, "beginning", "suspension", "ongoing" and "past" are only present in the "interventions" category).

Taking into account the aim of the project was the presentation of a timeline indexing the EHR, time expressions contained in the texts would also have to be classified, so we added a level-1 "time" node to the ontology that would allow the categorization of frequency, duration, date and other general temporal expressions.

## 3  Annotation and Machine Learning Model

Having had the ontology established, we needed a large human-labelled data set to feed the model training. Machine learning models usually require large volumes of training annotated data, so they can perform reliably and with high-accuracy levels. These large data collection efforts which require extensive human annotation can be very costly and time-consuming [8], so we needed to define a strategy that would be the most time and cost-saving.

### 3.1  Preparatory stage

We started by selecting the appropriate texts to be included in the training, validation and test datasets. Considering the project's goals, these were the types of EHR included in the datasets: daily clinical notes, test results, discharge summaries, medical histories. After determining the average length of each document and taking into account the ontology coverage, we decided on a 3000-document dataset, which was considered to be enough data for the model to perform with high accuracy after training (see **Table *1***).

**Table 1.** Dataset description.

| Set | vocabulary | pretrained vocabulary | sentences |
| --- | --- | --- | --- |
| Train | 51789 | 49782 | 73099 |
| Dev | 51789 | 49782 | 4216 |
| Test | 51789 | 49782 | 4018 |

Due to privacy concerns and data protection [9], we had to make sure all the documents included in the data set were anonymized and did not contain any details that could identify patients.

We then had to decide on a tool that would allow annotators to work on the documents. Annotation tools are essential for the development of natural language processing algorithms [10], so it was important to use one that would speed up the process, be intuitive and user-friendly, and allow the annotators to quickly select the phrase and the annotation category from the ontology tree.

Even though there are many open-source and commercial annotation tools available [10], like BRAT [11], we decided on building a new one that would cater to the needs of this specific project. This tool had to be web-based, to allow its access from a browser without having to perform any kind of software installation, making it possible to use across different operating systems and environments.

Included in this preparation stage of the annotation process was also the definition of the guidelines and the recruiting and training of a team of annotators. As this annotation task was domain specific and focused on categorizing medical knowledge in clinical text [12], the annotators had to be recruited among the medical community. This team was composed of 14 physicians of different specialties and medical students, who started working on the annotation under the supervision of the clinical team and according to the developed guidelines.

### 3.2 Annotation stage

The annotation tool allows for quick selection of spans and the subsequent presentation of the ontology, which enables annotators to select the right category with only a few clicks or by looking for the desired category in a search box. The modifiers are also presented and changed according to the category level. This prevents annotators from choosing modifiers that are not expected to be used in some categories (for instance, using the modifier "chronic" with the category "Exams/Tests").

A color scheme and pop-up tags were implemented in the annotation tool to easily distinguish between the distinct categories in the annotated texts. These features, as simple as they might seem, serve as great aids when reviewing the annotations and allow for the quick detection of errors (see **Figure *1***).

ANÁLISES :
Neutropenia ligeira ( sem leucocitose ) : 49 . 5 ; monocitose : 21 . 7 .
AST : OK ; ALT - 44 ; GGT : 29 .
Colesterol total : 216 ; colesterol HDL : 4    Achado em Análises
Restante normal , incluindo função tiroideia , vit . B12 , ácido fólico , ureia , ionograma e f . renal .

**Figure 1.** Example of an annotated document using the annotation tool.

A pane with the list of the annotated spans was added to the interface of an open document, so the annotators could review all the annotations and easily remove any annotated span if needed. Other useful features, such as the ability to annotate all the occurrences of a specific span in the same document, were added later while the annotation was in progress, in accordance with the suggestions of the annotators. This close collaboration between the clinical team, software developers, computational linguists and annotators helped us build a powerful and versatile annotation tool that was suitable for this project but also adaptable to any kind of annotation for future ones.

Soon after the annotation process began, and to validate the quality and coherence of the annotation, we took an annotated sample from the set and assessed the inter-annotator agreement. We did this in several iterations and worked with the team to fix the issues detected and raise the agreement metrics. We were aware that, for this kind of annotation, full agreement was an almost impossible task, but, since the quality of machine learning algorithms is dependent on the quality of the annotated data, and machine learning algorithms learn to make the same mistakes as human annotators, if these mistakes are made consistently [13], it was important to reach a high inter-annotator agreement. The annotators were encouraged to discuss among themselves any issues and doubts that would come up, and all the decisions were included in the guidelines for future reference.

Considering the ontology's coverage and the project's scope, almost all the information contained in the EHR selected was relevant, so we knew the annotation was going to be a strenuous task, as it would require annotating a substantial percentage of the documents' contents. This was confirmed by the annotators' feedback down the line while the annotation task was in progress. In comparison to other annotation tasks related to named entity recognition (NER), the annotation of EHR poses different challenges, namely the density of the annotated spans, particularly when it comes to annotating exam results, which are often large and detailed reports with long sentences.

The annotators took about 6 months to fully annotate the datasets, which, after going through the validation and uniformization tasks to detect discrepancies in the annotations, were then ready to be used for the training of the model.

### 3.3   Machine learning model

In the medical domain and other scientific areas, it is often important to recognize distinct levels of hierarchy in entity mentions, such as those related to specific symptoms or diseases associated with different anatomical regions. Unlike previous approaches, we built a transition-based parser that explicitly models an arbitrary number of hierarchical and nested mentions and proposes a loss that encourages correct predictions of higher-level mentions. We further proposed a set of modifier classes which introduces certain concepts that change the meaning of an entity, such as absence, or uncertainty about a given disease. Our model achieved state-of-the-art results in medical entity recognition datasets, using both nested and hierarchical mentions. The proposed model [14] was published at The 2nd Clinical Natural Language Processing Workshop (within NAACL 2019), Minneapolis, USA. Since the publication, the model has been improved by using contextual representations like BERT [15], further increasing the global NERC F1 results by 1.5%.

## 4   Integration

We developed an interface to present to the physician a set of antecedents based on relevant clinical information in the electronic clinical record, J One. This is expected to be most useful in the first contact with the patient. This way, we reduce the time the doctor spends looking for information about the patient, and we maximize the time spent with the patient.

APIs were created to return information from two NLP engines, the MATPD responsible for text processing at the lowest level (e.g., error removal, ambiguity, indexing, etc.) and the SAACE responsible for the temporal and semantic contextualization of entities of text with knowledge bases.

The development of these APIs was made taking into account a set of assumptions and methods, according to the consumption of information, which returned information about the annotation catalog updated at its various levels, about the list of all texts of a given patient and of all concepts marked in each text, return the count and index of all texts of a certain process number "that marked one or more concepts", or return the count and index of all texts of a certain number of process "that signaled one or several concepts".

From here, an interface was built with a graph of extracted mentions in which the size of the words is related to the number of times they are cited in the patient's EHR, a word cloud, allowing to grasp a good and immediate overview of the patient's history.

From the interaction with each word in the cloud, the clinician can easily view the citation history of the selected reference and read the specific text of a given citation if desired (see **Figure** *2*).

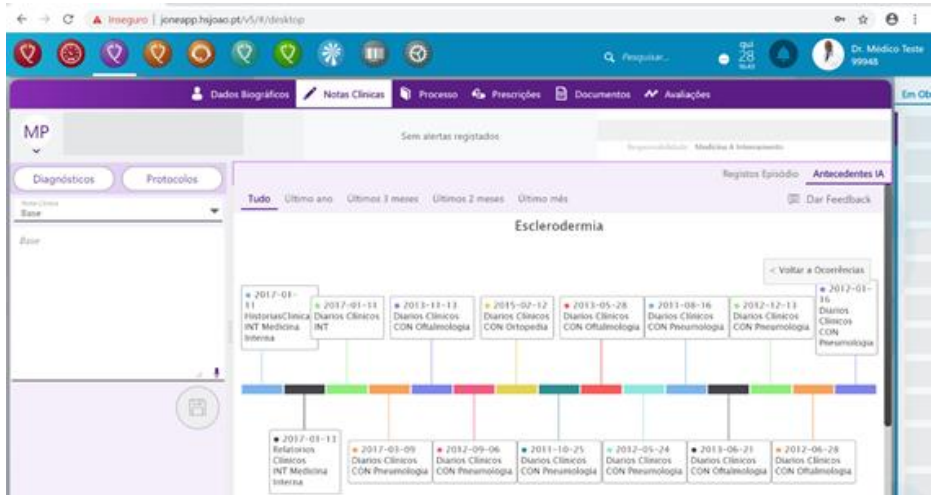

**Figure 2.** Timeline indexing the primary EHR texts where the reference "Esclerodermia" is cited, identified with date, type of record and specialty/department. By clicking on the item, the user is taken to the original text in the EHR.

We tested the system's usability by a group of end-users. The perceptions were very favorable in general and regarding visualization and performance (velocity, accuracy and yielding). The users pointed as positive aspects the rapid identification of the patients' history, the presentation of diseases by systems allowing an easy and grouped view, the usefulness of the ability to confirm specific diseases, the possibility of reading the original source document and locating events in time. Suggestions to improve included a better capacity of not distinguishing words that refer to the same construct (for example typographical errors, pharmacological and commercial names of drugs) and a clearer presentation of denials.

## 5      Future work

The consolidation and expansion of the work done so far includes several types of activities to answer different types of scientific/clinical questions, materializing other applications of the outputs of our NLP solution. This will most often involve the semantic contextualization of entities through the parametrization of the correspondence between our ontology and existing bases for specific types of constructs [16], such as the International Classification of Diseases [6] or the Orphanet Rare Disease Ontology (ORDO) for pathological entities [17], SNOMED CT for diagnoses, clinical findings and exams/devices [7], an international classification of drugs such as for example the World Health Organisation's ATC [18], the Medical Dictionary for Regulatory Activities (MedDRA) [19] for adverse reactions, and so on.

An application for routine care of individual patients intends to allow physicians to question the whole bulk of historic information directly, for instance: "Has the patient taken any antibiotic in the last 3 months? Which one?". When prescribing an antibiotic, it is relevant to consider whether the patient has taken antibiotics in the previous weeks. Having drug allergies should be confirmed before prescribing any drug. Previous hemorrhagic events contraindicate certain therapies that may be urgent. Disregarding this type of information allows for less-than-optimal decisions, with an important impact on patient safety.

An optimized access to the patients' records may also contribute to several activities in clinical research, such as feasibility assessments, finding eligible patients to clinical trials, identifying adverse events and outcomes.

From a managerial perspective, we intend to provide a supporting tool to the team that codes hospitalizations and surgeries according to the ICD-10 and the identification of patients eligible to specific financial programs such as comprehensive payment for some cancers.

We intend to validate and improve the capture of specific constructs, particularly clinical findings, including signs/symptoms and findings in exams that are usually described in text. These are essential for diagnosis and classification, for case definition, thus providing a basis to study diagnosis accuracy, to increase the sensitivity in identification of cases for several purposes and to obtain structured data suitable to include in quantitative analysis, if enough validity and precision are attained. This activity will focus on heart failure, leveraging previous work from our group, not only in terms of expertise in the subject but also in terms of availability of patient cohorts that were previously characterized in detail for scientific purposes. The syndromic nature of heart failure with heterogeneous presentation and much misdiagnosis supports the relevance of being able to use clinical information in text to classify patients. Similar approaches to other diseases and syndromes will certainly then emerge and make use of the richness of narrative clinical description of cases.